\documentclass[journal]{IEEEtran}
\usepackage{blindtext}
\usepackage{graphicx}

\usepackage[font=small]{caption}
\usepackage{amssymb}
\usepackage[cmex10]{amsmath}
\usepackage{float}
\usepackage{subcaption}
\usepackage{ tipa }
\usepackage{hyperref}
\usepackage[table]{xcolor}
\usepackage{mathpazo} 
\usepackage[final]{pdfpages}
\usepackage{csquotes} 
\usepackage{cite} 
\usepackage{siunitx} 
\usepackage{scrextend} 
\usepackage{courier} 
\usepackage{placeins} 
\usepackage{pgffor} 
\usepackage{csquotes}
\usepackage{tabularx}
\usepackage[english]{babel} 
\usepackage{blindtext}
\usepackage{array} 
\usepackage{mdwmath} 
\usepackage{eqparbox} 
\renewcommand{\thetable}{\arabic{table}}

\usepackage{enumitem}
\usepackage{flushend}

\usepackage{sidecap} 

\usepackage{hyperref} 

\begin{document}
%
\title{Reach Out and Help: Assisted Remote Collaboration through a Handheld Robot}
%
%
%

\author{Schachar Janis I. Stolzenwald and Walterio W. Mayol-Cuevas \thanks{ 
		Department of Computer Science, University of Bristol, UK,
		{\color{white}....} 
		janis.stolzenwald@bristol.ac.uk, wmayol@cs.bris.ac.uk
	}
}
\maketitle

\begin{abstract}
	  A remote collaboration setup allows an expert to instruct a remotely located novice user to help them with a task. Advanced solutions exist in the field of remote guidance and telemanipulation, however, we lack a device that combines these two aspects to grant the expert physical access to the work site. We explore a setup that involves three parties: a local worker, a remote helper and a handheld robot carried by the local worker. Using remote maintenance as an example task, the remote user assists the local user through diagnosis, guidance and physical interaction. We introduce a remote collaboration system with a handheld robot that provides task knowledge and enhanced motion and accuracy capabilities. We assess the proposed system in two different configurations: with and without the robot's assistance features enabled. We show that the handheld robot can mediate the helper's instructions and remote object interactions while the robot's semi-autonomous features improve task performance by 37\%, reduce the workload for the remote user and decrease required verbal communication bandwidth. Our results demonstrate the effectiveness of task delegation with the remote user choosing an action and the robot autonomously completing it at the local level. This study is a first attempt to evaluate how this new type of collaborative robot works in a remote assistance scenario. We believe that the proposed setup and task-delegation concepts help to leverage current constrains in telemanipulation systems.
\end{abstract}


%
\IEEEpeerreviewmaketitle

%
%
%
%

\begin{figure}[t]
	\centering
	\begin{subfigure}[b]{0.99\linewidth}
		\centering
		\includegraphics[width=0.99\linewidth]{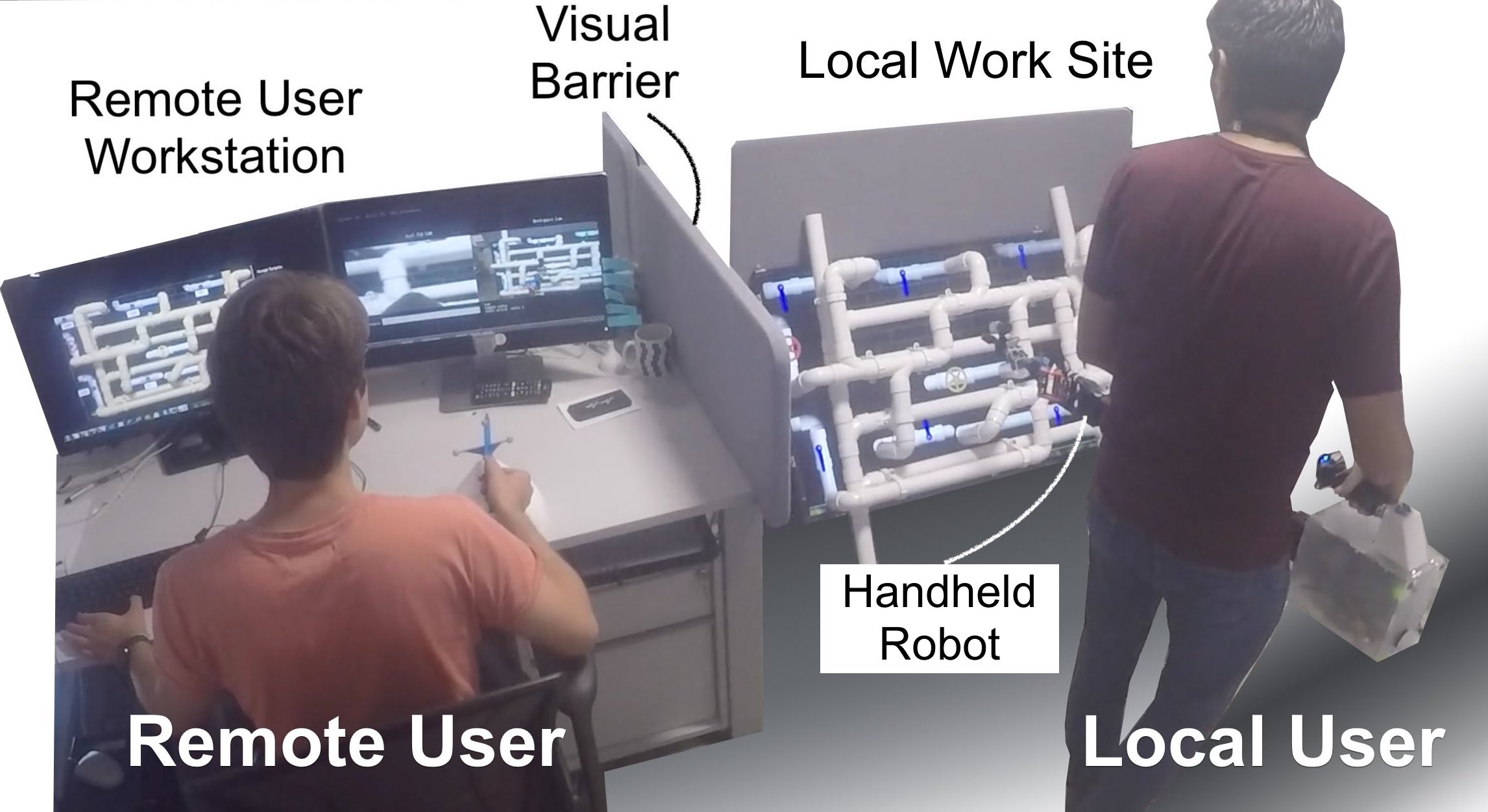}		
	\end{subfigure}
	\hfill
	\begin{subfigure}[b]{0.99\linewidth}
		\vspace{1em}
		\centering
		
		\includegraphics[width=0.99\linewidth]{"Pictures/Remote_Setup"}
	\end{subfigure}
	\caption{\textbf{Overview of the Remote Assistance Experiment Setup.} The remote assistance system allows a remotely located user to assist a user at the local work site through a remote-controlled semi-autonomous robot, which allows for inspection and physical interaction. Using the remote workstation, the local workspace can be perceived through robot-equipped cameras. The local user helps with reaching through moving the robot to specific locations in the scene as they follow the remote user's instructions. 
}
		\label{fig:workspaceoverview}
\end{figure}
\section{Introduction}

\IEEEPARstart{T}{his work} introduces an assistive collaboration system that allows a remote expert to guide a local user through a task using a handheld robot, which is carried by the local user. The aim of this research is to explore how the robot could work as a mediator between the two human parties based on the concept that the expert could perceive and explore the work environment through the robot and then delegate task operations to the robot that in turn completes them in part autonomously. The proposed system is tested using the maintenance of a plumbing system as an example task, which covers a broad variety of common human-robot-human interaction problems. We show that the robot's autonomous contributions to the task leverage a workflow  that is characterised by efficient interaction.

Collaborative remote assistance tasks usually involve a less experienced person (local worker) who has to manipulate a set of physical objects with the help of a remotely situated expert. In the context of a complex problem, the local worker has limited knowledge about required operations and relies on the instructions of an expert. 
Examples for such tasks are maintenance \cite{Gauglitz:2014hr} and inspection \cite{Linn:ZL0DUdp_} of remotely located systems and expert-guided surgery to train remote novice surgeons \cite{Ballantyne:2002jc,Dell:2003tc}. Current systems solve collaboration challenges such as spacial referencing and communication, however, they limit the expert's view to a camera that is either stationary or worn by the local user \cite{Gauglitz:2014gr}. 
What is missing is a system that allows the remote user to explore the environment for inspection and grants physical access to decouple the solution from the local worker's competences. 

A new approach to remote collaboration is to use a robot to enable a remotely located user to help with a task in the local user's workspace. For example Veronneau et al. \cite{Veronneau:2020dr} present a remotely actuated waist-mounted robot that can assist in various tasks through the help of a remote assistant. With these recent developments, the question raises to what extent a robot's partial autonomy and task knowledge could be used to leverage remote interaction. 
Answering this question is particularly relevant for more complex tasks, i.e. in tasks where the solution is not immediately obvious to the local user e.g. due to a lack of task expertise. With the robot as a partially autonomous assistant, such a setup would go beyond a master-slave relationship and more towards a working together paradigm with the users and the robot each contributing to the task.


Remote maintenance is of particular interest for industrial applications \cite{Bordegoni:iANORb1F}. Modern products and plants are characterised by increasing complexity which requires high expertise to diagnose and solve problems. However, it might be expensive to get an expert on site and relying on the consultation of a manual is inefficient \cite{Wang:2014bt}.
Therefore, the development of remote assistance systems has caught researchers' attention in recent years. 
Some solutions have considered remote guidance through augmented reality (AR) \cite{Bordegoni:iANORb1F} and semi-autonomous telemanipulation systems \cite{Marturi:2017cm}, i.e. a system that solves some subtasks such as grasping autonomously. However, what is missing is a remote assistance setup that combines the advantages of physical access through telemanipulation with the ones of cooperative guidance and task solving.

With humans and robots working together in complex tasks, defining the collaboration style in a scenario that is characterised by mixed-initiative interaction is of particular importance to enable each party to contribute to the task with their respective strengths at suitable times \cite{Hearst:1999vr}. Traded control is a subdomain of mixed-initiative interaction, where a human is in charge of controlling the robot for some parts of the tasks, while the robot is supervised but autonomous during other parts of the task \cite{Kortenkamp:1997tc}. This concept could benefit the interaction with remote maintenance robots: The remote expert could control the robot during the inspection part, in which they explore the scene, while the robot could take over control once the expert makes a diagnosis and decides on what needs to be done. Regarding the proposed pipe maintenance test case, examples for such semi-autonomous operations could be to adjust valves in the system or to inspect gauges and pipe elements. What is necessary for the autonomous execution is the robot's knowledge about the task, i.e. the location and identity of task objects and how to interact with them. For example, adjusting a valve is a different form of interaction than checking a pipe element for cracks.

Handheld robots \cite{GreggSmith:2015bh,GreggSmith:2016cz,GreggSmith:2016hn,Stolzenwald:2018un,Stolzenwald:2019wi, Elsdon:2017is,Elsdon:2018kl} are intelligent tools that can process task knowledge and environment information for semi-autonomous assistance in collaborative task solving. Notably, this type of robots combine these robotic abilities with the natural competences of human users for negotiating obstacles and resolving complex motion planning tasks effortlessly. We argue, that such a system could bridge the aforementioned gap between remote guidance and telemanipulation, with the handheld robot helping both the effective communication between the workers and task outcomes. The proposed system allows for spacial referencing through tooltip pointing, enables the remote user to perceive the environment through a steerable camera and supports remote physical interaction through the tooltip interface and control. 

The spacial coordination between the remote user, the robot and the local user is a challenging problem. We see the robot's accuracy and motion capabilities as a potential to delegate small-scale tasks to the robot. That way, remote users can focus on strategies and their communication to complete the task rather than wasting their capacity on fine-tuned tool tip motion required for the task execution. 

%
%
%

We explore the proposed remote assistance setup guided by the following research question:    

\begin{enumerate}[label=\textbf{Q}]
	\item How does the handheld robot's autonomy and task knowledge affect performance and communication in a remote assistance setup?
	\label{Q}
\end{enumerate}

This work involves  an extended version of the open robot platform, introduced in \cite{GreggSmith:2016cz}, which is coupled to the proposed workstation for telepresence access. 
As the interaction between the handheld robot and its carrier has been investigated extensively in previous studies \cite{GreggSmith:2016hn,GreggSmith:2015bh,Stolzenwald:2019wi,Stolzenwald:2018un}, the focus of this work lies on the experience of the remote access i.e. the role of the remote expert. We investigate the subjects' interaction in this human-robot-human setup and assess proposed semi-autonomous assistive features of the robot. An experiment overview can be seen in Figure \ref{fig:workspaceoverview}. 
The outcome of the studies are summaries in the following list of contributions:

\begin{itemize}
	\item As the main contribution, a  new paradigm for remote assistance through handheld robot collaboration is introduced. It allows for remote-teamwork with the handheld robot as a mediator of instructions and physical interaction.

	\item Another key contribution is the proposed concept of local task delegation through the robot's assistive features. We show that their use improves teamwork productivity and usability, hence why we suggest that they could benefit other telepresence systems too.
	
	\item The qualitative analysis reveals a repeated sequence of exploration, guidance, local task solving and retraction as a common emerging strategy in this novel collaboration setup. This knowledge is useful to guide future developments of similar human-robot-human setups.
\end{itemize}

A demonstration of the remote assistance system and examples for the robot's semi-autonomous assistance features can be seen in the supplementary video material.




\section{Related Work}
In this Section we review the current state of handheld robot technology and discuss challenges in the fields of remote guidance and telemanipulation.

\subsection{Handheld Robots}
The concept of a {\it general-purpose} handheld robot was first introduced by Gregg-Smith and Mayol-Cuevas \cite{GreggSmith:2015bh}, who propose an intelligent tool that can perceive the environment and is able to perform actions through an actuated tooltip, that are based on the robot's task knowledge such as subtask progress and current states. They prototyped a first version of the device with 4 DoF, which was used to study how it could be used for human-robot collaborative task solving. Their work demonstrates that the aforementioned features contribute to decreased workload as it provides guidance through pointing gestures and refuses commands that would lead to mistakes. 
This goes in line with findings from other works on intelligent tools that use task knowledge and shared control to assist in welding \cite{Echtler:2003uo} or painting \cite{Elsdon:2018kl}.

Further work in this domain introduced a handheld robot with a 6-DoF kinematics design \cite{GreggSmith:2016cz}, for the investigation of robot-human communication for user guidance \cite{GreggSmith:2016hn} and the perception of users' attention and intention for improved cooperation \cite{Stolzenwald:2018un,Stolzenwald:2019wi}.

These works aim to investigate how handheld robots and humans benefit from each others' strengths within a single-user collaborative setup. The developed concepts built on Moravec's paradox \cite{Moravec:1988un} but turn it into a collaborative benefit: Users carry out the broader tactical motion with their intuitive navigation and obstacle avoidance skills and benefit from the robot's speed, accuracy and task knowledge. The result helps to reduce the time of task execution and the number of errors and particularly helps users of handheld robots to carry out tasks in which they have limited expertise.

While the introduction of the robot's task knowledge brings advantages, a crucial aspect remains open, i.e. where this knowledge might come from, e.g. whether it could be learned \cite{AbiFarraj:2017ex} or derived from a remote expert and mediated through the robot \cite{Kritzler:2016ie}. In this work, we start with exploring to what extent a handheld robot is suitable for a remote assistance human-robot-human setup and whether the benefits of the robot's partial autonomy can be observed analogous to the aforementioned single-user instances. 

\subsection{Remote Guidance}
This work brings together the concepts of handheld robotics and remote assistance. The latter involves two key aspects i.e. remote guidance and telemanipulation.
Remote guidance systems allow a remotely located person to assist a local person through instructions and directions. Research in this field aims to overcome the limitations of traditional consulting methods such as audio or video calls \cite{Kritzler:2016ie}. 
Current solutions to communicate the local conditions of the task state require a camera at the worker's site that is either stationary
\cite{Wang:2014bt,Linn:ZL0DUdp_,Dell:2003tc}, portable (e.g. smartphone or tablet) \cite{Gauglitz:2014hr,Gauglitz:2014gr,Sodhi:JJe-ACW9,Bordegoni:iANORb1F}, worn by the local user 
\cite{KostiaRobert:2013un,Bottecchia:2010co} or operated by the helper \cite{Kuzuoka:DjdzJkGx,Anonymous:pTXsTX3C,Kurata:qhDyKTAw}.

These solutions have in common that the video feed of the local workspace is displayed to the remote helper either on a desktop screen, tablet or smartphone alongside audio communication. However, the methods are distinct in their respective communication features for the helper's instructions to the local user. For example, early work introduced video markups such as on-screen drawings and predefined shapes \cite{Dell:2003tc,Wang:2014bt}, which were later taken to a mixed-reality approach through 
AR technology \cite{Linn:ZL0DUdp_,Kangas:2018ky}.
Other solutions allow the helper to give instructions via hand gestures \cite{KostiaRobert:2013un,Sodhi:JJe-ACW9} or captured full-body motion \cite{Yamamoto:2018gt}. 
By comparison, less effort was spent on the exploration through embodied guidance i.e. mediated through a robot. An example is \textit{GestureMan} \cite{Anonymous:pTXsTX3C}, which is a mobile robot equipped with an actuated camera and pointing stick. However, the pointing mechanism is limited by a few DoF and thus not suitable for manipulation and dexterous gestures. 

We note that while the above-mentioned works offer efficient solutions for the communication of instructions, they do not allow any direct physical interaction within a collaborative setup. This leaves local workers in charge of carrying out any manipulative actions by themselves.

\subsection{Telemanipulation}
In contrast to remote guidance systems, telemanipulation allows a remote operator to execute physical operations i.e. through a remote-controlled robot rather than instructing another person. Application examples exist in the form of sedentary robots for the remote maintenance and inspection of machinery \cite{Bellamine:2004gd,Bellamine:2002kt}, telerobotic surgical systems \cite{Ballantyne:2002jc}. Furthermore, remote manipulation is used for safe physical interaction in hazardous environments e.g. the outer space \cite{Schiele:2003tk} in nuclear decommissioning \cite{AbiFarraj:2016ec}.
In terms of mobile robots for telemanipulation, the most advanced examples have been explored in the context of disaster response \cite{Haynes:2017jb} and the exploration of the outer space \cite{Imaida:2004bm} and deep sea \cite{Wedler:2018vn}. These systems enable a remote user to navigate through unstructured environments and manipulate physical objects. 

The existing solutions for telemanipulation are useful for environments that are inaccessible for humans and for scenarios with highly specialised robots and operators (i.e. surgery). However, the research question of how remote guidance and telemanipulation could be combined in a collaborative setup remains unanswered.

\section{Remote Assistance Study} 
In this study, we propose and test a remote assistance system with a handheld robot as its core research object. The collaborative setup includes three parties that are working together: one user at the local worksite, a remotely located expert user and a robot that mediates task information and the expert's instructions (Figure \ref{fig:workspaceoverview}). A pipe system maintenance task is used to assess the robot's assistive features with regards to the human-robot-human interactions, emerging in this setup. The experiment involves inspection and repairing of a piping system. Through the robot, the expert can check the pipes for hidden cracks and adjust a set of valves to regulate fluid currents.

\begin{figure}
	\centering
	\includegraphics[width=0.99\linewidth]{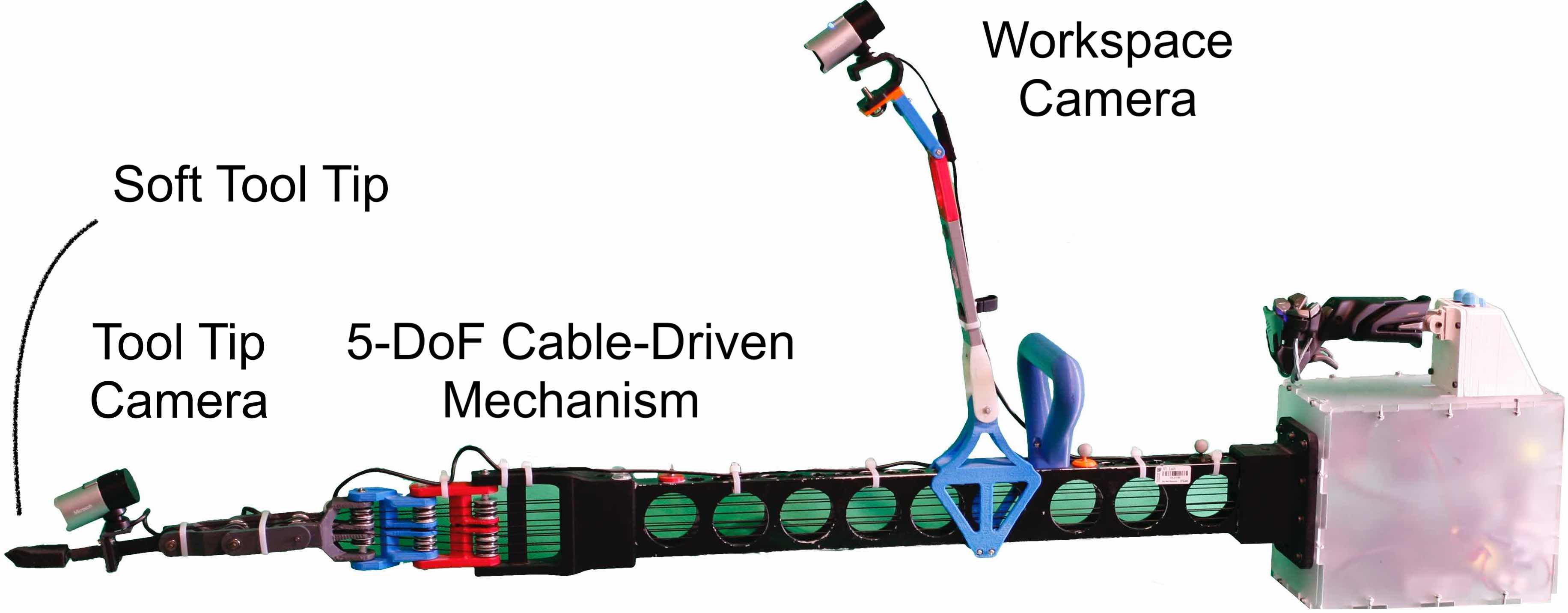}
	\caption{\textbf{Handheld Robot in the Remote Setup}. It has a remote-controlled tooltip with 5 DoF, equipped cameras and a tooltip for sensor simulation. The workspace camera grants the remote user a broad overview of the scene, while the camera at the tooltip allows for close observation of tooltip operations.}
	\label{fig:handheld-robot}
\end{figure}

The local user has physical access to the maintenance object but limited knowledge about the steps required to carry out the maintenance task. This knowledge is held by the remote expert, who in turn has no direct physical access to the worksite. To bring (local) physical access and (remote) task knowledge together, a camera-equipped handheld robot with 5-DoF motion capabilities (displayed in Figure \ref{fig:handheld-robot}) is held by the local user but controlled by the expert through the proposed remote interface for inspection, manipulation and gesturing. The handheld robot can assist in the task through semi-autonomous control features for subtask solving. For example, it can check selected pipes for cracks or change a valve to a value specified by the expert. These features are purposed to lower the remote user's cognitive load that would otherwise be required for the control of the robot, which allows for a shift of the expert's focus on high-level task planning. 


\subsection{Study Design}
For our experiments we use a within-participant design to compare the performance of the remote user and local user pairs using our proposed remote assistance system in two different conditions:

\subsubsection{Non-Assisted} 
The robot is steered manually by the remote user to guide the local user to task objects. After reaching an object, the remote user completes the task manually. Some tasks require verbal request of information about the system’s current state from the local user.
\subsubsection{Assisted} Initially, the robot is steered manually for guidance. When approaching a task object, the remote user can select it to delegate the interaction to the robot. The robot assists through locally fulfilling the task within its workspace and takes into account the system’s current state (detailed description in section \ref{sec:remoteAssistance}). 

The setup is a semi-simulated pipe system, which we use as an example for a collaborative maintenance task. Solutions for solving this task requires elements of common real-world assistance problems, such as inspection, diagnosing, instructing and manipulation.

\subsection{Hypotheses}
\label{sec:hypotheses}
Concerning the effect of the robot's assistance features on performance and collaboration, we hypothesise that with those features enabled:

\begin{enumerate}[label=\textbf{H\arabic*}]
	\item The time to complete the task would be reduced as the robot's assistance saves time.
	\label{H1}
	\item The robot's task knowledge and autonomy reduce the required total amount of communication and reduce the share of the local user.
	\label{H2}
	\item The perceived workload of both users would decrease as it gets transferred to the robot.
	\label{H3}
	\item The users' rating of the system's usability would be increased.
	\label{H4}
\end{enumerate}

\ref{H1} goes under the assumption that the robot performs local tasks faster when solving them autonomously, due to its accuracy and precise timing, i.e. the time for aiming and fine-tuning of motion is shorter and the robot can react faster. Plus, the operator can already start to plan a subsequent step, while the robot completes a previous step. Concerning \ref{H2}, it is expectable, that communication of both parties gets reduced concerning the amount that is used for coordination as the robot's autonomy helps to regulate spacial inaccuracies. Plus, in autonomous mode, the robot accesses information about the workspace site that is otherwise exclusive to the local user. Hence, a decrease in the local user's share of verbal communication is expected. As a result, executing the task becomes easier (\ref{H3}) and the robot more user friendly (\ref{H4}).

\subsection{Participants}
24 participants (all male, $m_{age} = 26.4$, $SD = 4.3$) were recruited. Most were students from our Computer Science Department, however, technical knowledge was not required. There was no financial benefit for their voluntary participation. They were evenly split into two groups for the role of the remote user and the local user, i.e to form 12 distinct pairs. 

\subsection{Collaborative Setup}
The task setup consists of two main areas, the local workspace site and the remote workstation (see Figure \ref{fig:workspaceoverview}). For the experiment, remote and local users were located in the same room, however, a visual barrier prevented them from direct interaction. They were allowed to speak to each other as if they were on a phone call. 

For the experiment, we used the handheld robot reported in \cite{GreggSmith:2016cz}, of which the mechanical design is publicly available \cite{Anonymous:XQistSCe}. The robot was adapted to  suit the requirements for a remote control setup. Inspired by examples from the field of remote assistance \cite{Wilson:2014el}, we followed a two-camera design to combine a detailed view with a more distant scene view.
As can be seen in Figure \ref{fig:handheld-robot}, the robot features a 5-DoF actuated tip and two cameras. The first camera is fixed to the robot's frame and delivers the overview of the current workspace. The second camera is positioned close to the tooltip so that it can be directed for exploration, whilst allowing a detailed view on tooltip operations. 

The remote user sat at a desk equipped with the remote interface, which allows for the perception of the robots' workspace and features a 5-DoF input for the remote control, as well as information about the system and its required goal states. A detailed description of the workstation is presented in section \ref{sec:remote_workstation}. The local user is located in the workspace where the physical task has to be completed. The user holds the robot in place for inspection and helps the remote user to reach objects for manipulation and diagnosis. The collaborative setup is shown in Figure \ref{fig:workspaceoverview}, additionally, a task demonstration is included in the supplementary video.

\subsection{Experiment Design}\label{sec:remote_experiment-design}

\begin{figure}[h]
	\centering
	\begin{subfigure}[b]{0.99\linewidth}
		\includegraphics[width = \linewidth]{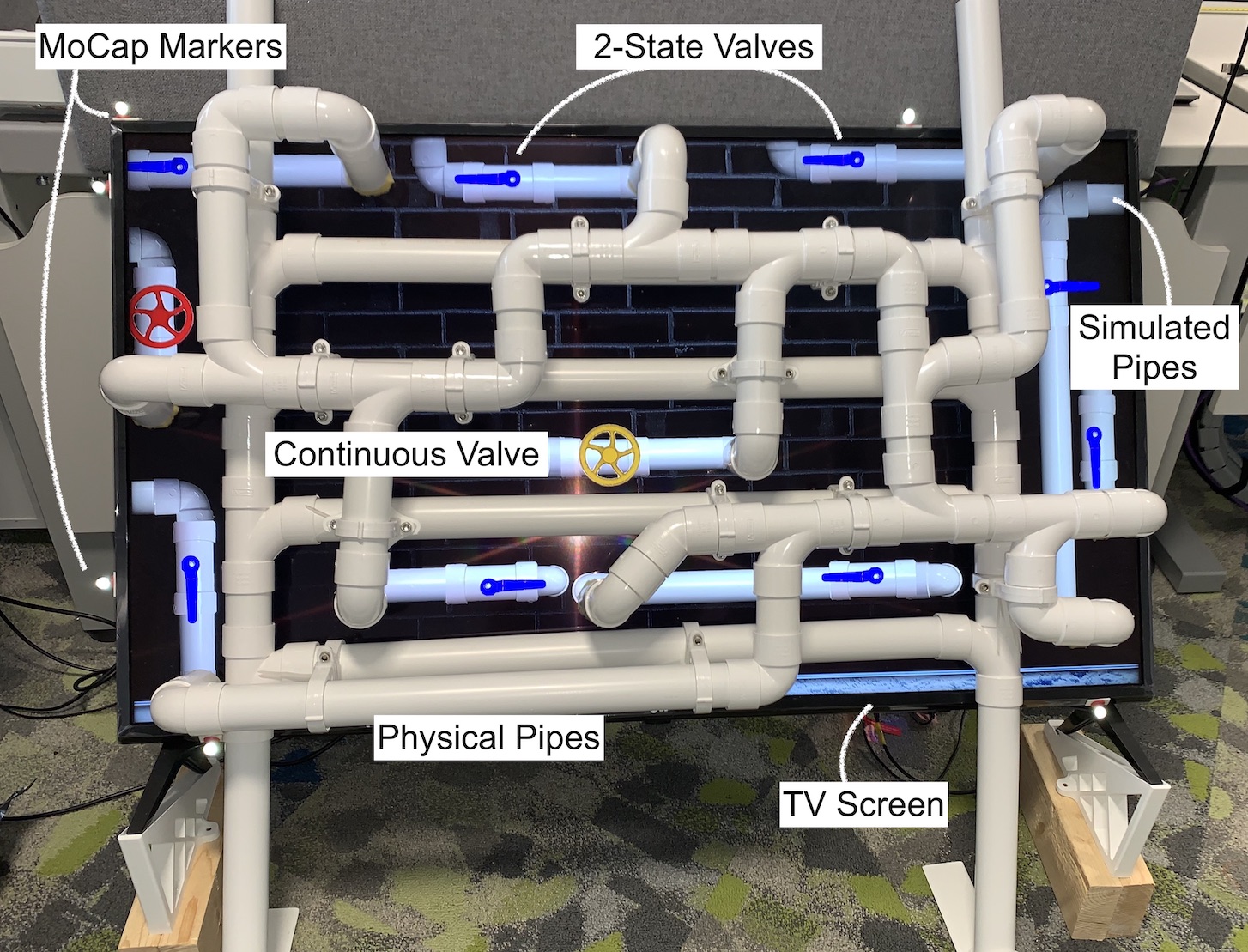}
	\end{subfigure}
	\hfill
	\begin{subfigure}[b]{0.99\linewidth}
	\vspace{0.5em}
		\includegraphics[width=\linewidth]{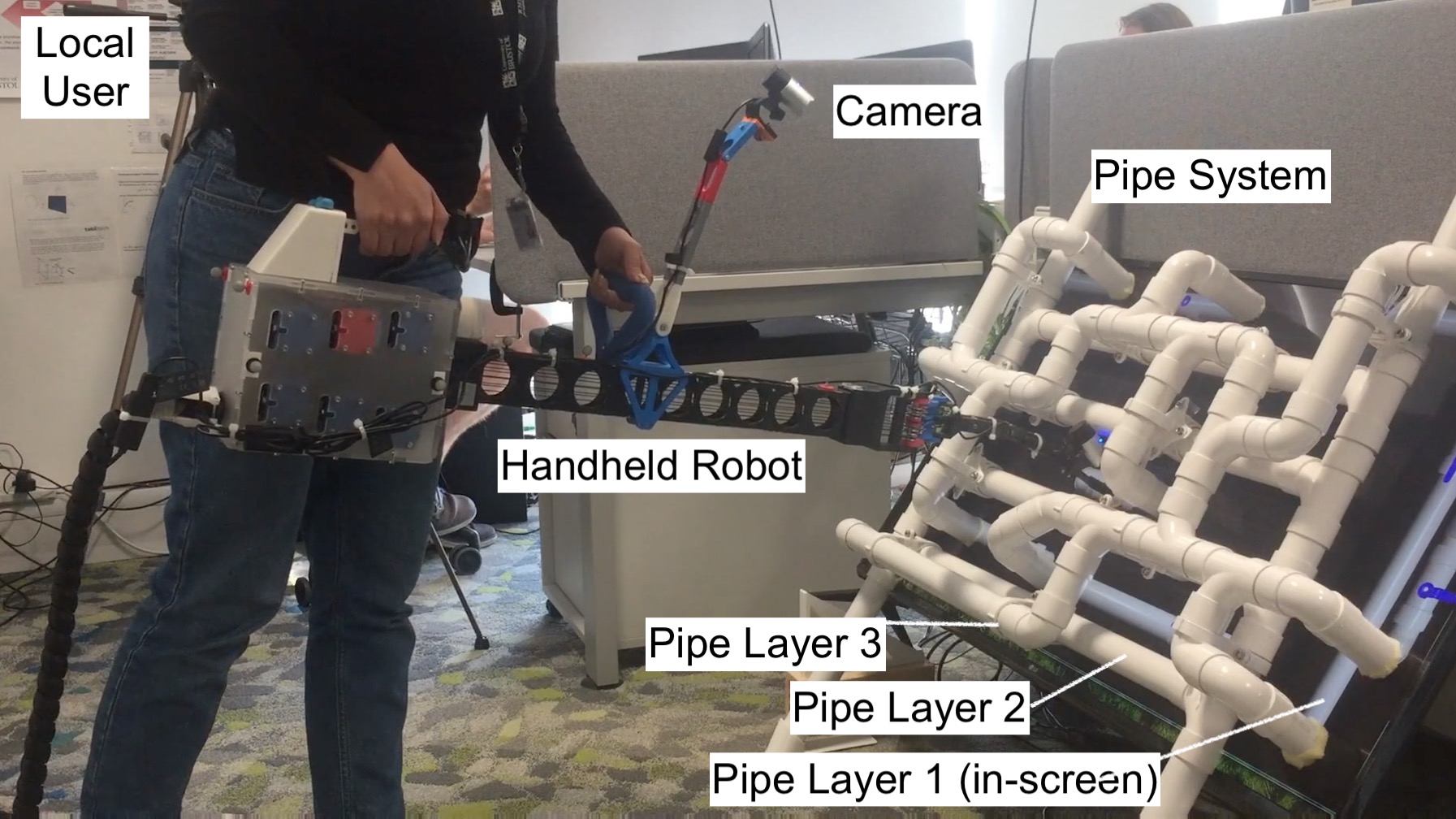}
	\end{subfigure}
	\caption{\textbf{Overview of the Mockup Pipe System}.
		The pipe system was used for the maintenance experiment task. Its main elements are the (blue) two-state valves, the (yellow and red) continuous wheel valves and the pipes. The piping is organised in three layers, one is simulated in the LCD TV screen (top) and the other two layers made of physical pipes in front of it (bottom). A top view of the experimental layout is presented in Figure \ref{fig:workspaceoverview}.}
	\label{fig:pipesystem}
\end{figure}

To test the proposed hypothesis \ref{H1} to \ref{H4} (cf. Section \ref{sec:hypotheses}), the experiment setup requires the following elements:

\begin{itemize}
	\item Common remote collaboration tasks such as inspection, diagnosis, remote guidance and manipulation.
	\item  Some information about the task goal should be exclusive to the remote user. This is their expertise in the system.
	\item Some information about the current state of the local work site should be exclusive to the local user. This is the worksite specific knowledge.
\end{itemize}

Following the above criteria, a semi-simulated maintenance task of a pipe system was chosen. The task is simple enough to fit the scope of a user experiment and allows adding cognitive load to test more complex scenarios.
The setup consists of a network of pipes, valves and gauges. While the majority of the pipes are a physical system, the valves and gauges are simulated through a display in the background of the pipe system (Figure \ref{fig:pipesystem}). The gauges were placed so that the remote user cannot look at the gauges while carrying out work on the pipes i.e. information about their current state is exclusive to the local user. The values of the gauges depend on the state of the valves. 
There are two different kinds of valves, the first kind has only two discrete states i.e. open and closed while the wheel-shaped valves are continuous. Each valve has a specific contribution to a gauge in the open state. The system contains 8 discrete valves, 2 continuous valves and 3 gauges. 


%

The experiment consists of two main tasks, adjusting the valves and checking the pipes for cracks. 
The remote user holds knowledge about the target values of the gauges and the contributions of the respective valves and consequently knows what changes need to be done. However, initially, the remote user does not know the pipe system's current state, i.e. valve states and readings of the gauges. Retrieving this information requires either a visual exploration of the work scene or verbal requests. 

When the soft tooltip of the robot touches the valve, the remote user can press an activation key to turn it open or close. This manipulation is simulated through a 2D animation of the valve handle/wheel in the screen and the associated gauge value changes accordingly. For simulation purposes, the touch of the robot's tip is registered using motion capturing\footnote{OptiTrack: optitrack.com}, which enables 3D localisation of the handheld robot and the screen surface. 

For the pipe checking task, a sonar sensor is simulated. The procedure of taking a measurement is inspired by \cite{Bellamine:2002kt} where a sensor tip is placed on a machine part for a short duration to check the condition of the material e.g. for crack detection. Similarly, here, the robot's tip needs to be in contact with a pipe to be checked for a few seconds while the remote user activates the sensor reading. 

There is no predefined order in which valves have to be opened or closed and pipes to be checked. In that way, the remote user has to come up with an individual strategy for a solution, which brings the task closer to real-world problems. The maintenance task is completed when all gauges display the desired target values and a predefined set of pipes is successfully checked with the sensor. 

The task setup stimulates cooperation between the users since neither party could solve the task on their own. The local user lacks knowledge about the specific task goals and relies on the remote expert's guidance, who would not be able to access the workspace without the help of the local user.

\subsection{Remote Workstation}
\label{sec:remote_workstation}
\begin{figure}[h]
	\centering
	\includegraphics[width=0.99\linewidth]{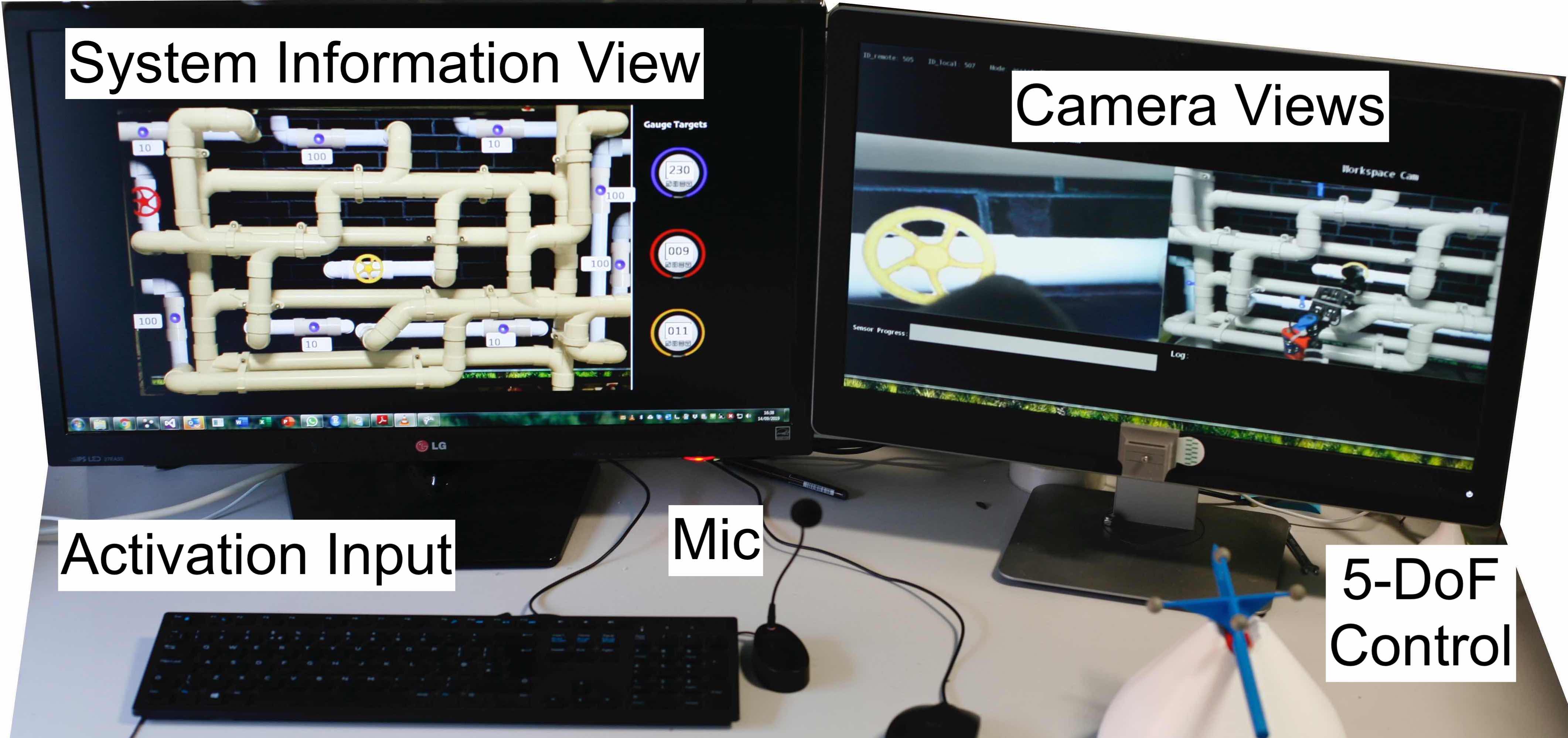}
	\caption{\textbf{Remote Workstation.} The remote user has access to specific system information (see system information view) and can see the local work site through the camera views. The robot is remote controlled using the activation input and the 5-DoF control. The position of the remote work station with respect to the overall experiment setup is presented in Figure \ref{fig:workspaceoverview}.}
	\label{fig:remoteworkstation}
\end{figure}

The workstation of the remote user consists of three main units: the robot control system, a display with the robot states and another one containing task system information. This design of the visual interface is in essence derived from solutions reported in several remote assistance studies e.g. \cite{Bordegoni:iANORb1F,Gauglitz:2014gr}, while the positioning of the cameras and the spacial input is inspired by previous work on remote manipulation\cite{Skilton:2018gw}. An overview can be seen in Figure \ref{fig:remoteworkstation}.

\begin{figure}
	\centering
	\includegraphics[width=0.99\linewidth]{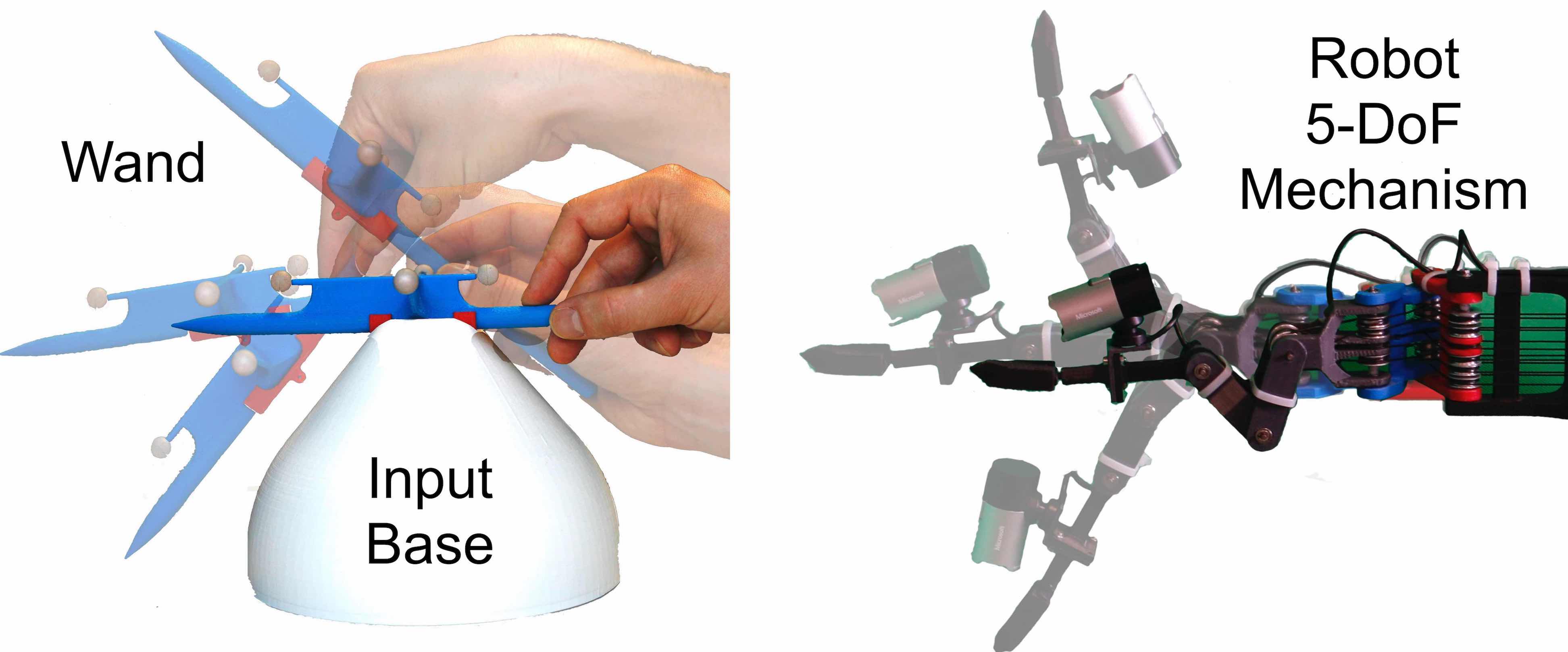}
	\caption{\textbf{Remote Control of the Handheld Robot}. Example of a remote user’s 5-DoF spacial input and the robot response (transparent images). The non-transparent position is the home position, which is centred within the robot’s work space.}
	\label{fig:5dofinput}
\end{figure}

The main part of the control unit is the 5-DoF input, which is realised through a wand, which is tracked through motion capturing. Its relative position and pose to the base socket is replicated by the robot arm with respect to its local reference frame. To account for the limits of the robot's workspace, the wand is attached to the base. 
The initial position allows the remote user to either reach out or retreat as demonstrated in Figure \ref{fig:5dofinput}. Tooltip operations such as manipulation and sensor activation can be triggered by pressing the space bar of a keyboard. 

The view of the robot state contains a split view for the two cameras, a progress bar for the simulated sensor state and a log protocol of robot actions. The screen with the system information shows an overview of the piping routing with valve locations, required values for the respective gauges and indications for which pipes need to be checked. It does not display the current state of the valves. 

As the robot is wired to the remote interface, the proposed testing setup does not take into account possible lags of the camera or tip actuation which might affect collaboration in a long-distance setup. 

\subsection{Robot Assistance Condition} \label{sec:remoteAssistance}

When the robot is in the assisted condition, a set of features are enabled which incorporate task knowledge and navigation capabilities. In this state, the remote user is no longer required to complete detail motion. Instead, he/she selects an object to interact with and the robot aims for it when he/she presses the activation key. For example, the remote user could roughly direct the local user to a valve, activate the assistant and the robot completes the manipulation. The robot has knowledge about the gauges target values and can, for example, turn open the continuous valves until the associated gauge matches the required value. Similarly, the robot helps with a world-stabilised positioning of the sensor on the pipes' surface during crack detection and retreats when the measurement is complete.

\subsection{Trial Procedure and Data Collection}
\label{sec:trial}
Before the start of the experiment, the participants were given an introduction to the system. Both robot conditions were practised with the experimenter taking over the role of the matching team member until the participant felt confident.
Within the pairs, the experiment task was executed one time for each of the conditions \textit{non-assisted} and \textit{assisted}, which were counterbalanced across the trials. This is to cancel out within-pair training effects and preferences. The roles were never swapped and local users were never matched with another remote user to avoid additional hierarchical dependencies in the experiment data. The initial state of the valves, the gauges' target values and which pipes would require checking were randomised in such a way that the number of actions required to solve the task remained constant for each trial. 
Participants were asked to complete the task swiftly and were informed that their completion time was recorded to measure their performance.

Investigating the proposed research question \ref{Q} requires metrics for the effectiveness and quality of human-robot interaction. As the focus of this work is on collaboration rather than solving the specific task, using a resulting product of a task as a measure of performance (e.g., traditionally used to assess the performance of surgical robotics \cite{Troccaz:1996ck}) is unsuitable here.	In accordance with the evaluation criteria established by Nielsen et al. \cite{Nielsen:2008vu}, we use task completion time as a measure of performance and perceived task load and usability as measures of users' involvement. With this in mind, the completion of the experiment task was followed by a record of the NASA Task Load Index (TLX) \cite{Hart:1988ho} and a System Usability Score questionnaire (SUS) \cite{brooke1996quick} for both subjects and for each condition. 

The TLX \cite{Hart:1988hoa} is a standardised test that measures an individual's perceived task load based of an average rating of the six aspects: Mental Demands, Physical Demands, Temporal Demands, Own Performance, Effort and Frustration. The test quantifies perceived task load with a score from 0 to 100, i.e. low to high (lower is better), respectively. The NASA-TLX has been in use for over 30 years now and it has already proven useful in previous handheld robot studies, e.g. in \cite{GreggSmith:2016hn,GreggSmith:2015bh}.

The SUS \cite{brooke1996quick} is a technology-independent rating based on a 10 Likert scale questions focusing on an individual's experience during usage. The test quantifies usability on a scale from 0 (low) to 100 (high), where a higher value is better than a lower one. The test has been in place since the late 90s and has been used to assess telerobotic systems before \cite{Adamides:2017jf}. Therefore, it using it in this study is a natural choice.

Additional experiment metrics recorded were the time to complete the task as well as voice recordings from microphones, which were placed close to each user. 
The audio material was later transcribed to derive word counts for the analysis. 
As an estimate of prior experience with video games, all subjects were asked for the average weekly amount of hours they usually spend on gaming. Plus, video recordings of the trials were taken for a qualitative assessment.

Furthermore, the required time to complete the task was recorded as well as voice recordings from microphones, which were placed close to the respective users. 
The audio material was later transcribed to derive word counts for the analysis. 
For an estimate of prior experience with video games, all subjects were asked for their average weekly amount of hours they usually spend on gaming. Furthermore, video recordings of the trials were taken for a qualitative assessment.

The experiment series was completed with: 12 pairs $\times$ 2 conditions $=$ 24 data points. 

\section{Results}
\label{sec:remote_results}

The analysis of the experiment data is divided into two parts, a quantitative and a qualitative assessment.

\subsection{Quantitative Analysis}
To assess the effect of the robot's condition on task performance and collaboration, we compare completion time and dialogues' word counts as well as TLX and SUS results between the two condition groups. Concerning these metrics, a series of paired \textit{t}-tests was performed with the robot's condition as an independent variable. The results are summarised in Table \ref{tab:remote-expert-ttest-results} and illustrated in the diagrams of Figure \ref{fig:resultDiagram}.

\begin{table*}[h]
	\centering
	\includegraphics[width=0.95\linewidth]{"Pictures/Remote_Expert_ttest_Results"}
	\caption
	{ \textbf{Summary of Quantitative Analysis.} \textit{t}-test results for the analysis of differences in average means for completion time, word count, TLX and SUS scale depending on whether the robot's assistive features were enabled. Starred values indicate a significant difference. The distribution of the data can be seen in the associated diagrams in Figure \ref{fig:resultDiagram}. Levels of Significance: ns (not significant) $p \geq .050$; * $p < .050$; ** $p < .010$; *** $p < .001$.}
	\label{tab:remote-expert-ttest-results}
\end{table*}

\def \myTextwidth {0.45}
\def \myScale {0.9}
\begin{figure*}
	\centering
	\begin{subfigure}[b]{\myTextwidth\textwidth}
		\centering
		\includegraphics[width=\myScale\linewidth]{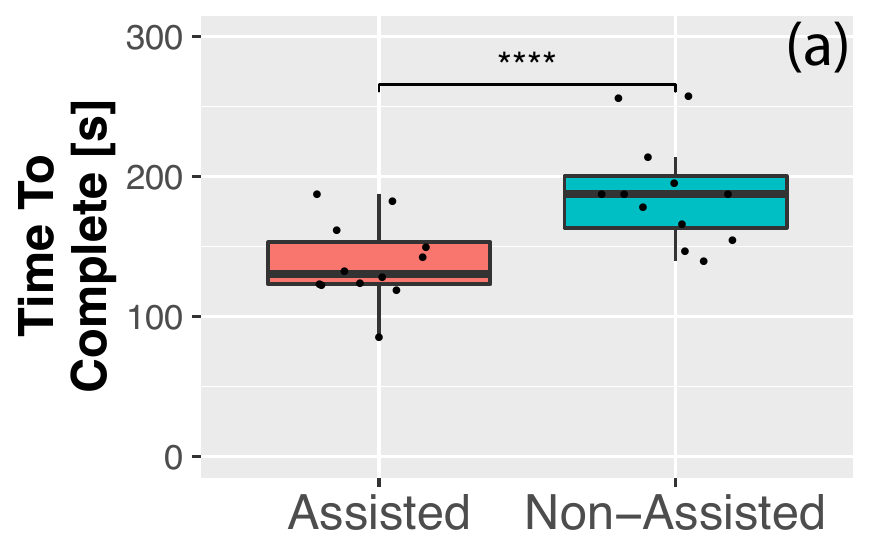}
		\label{fig:timetocomplete}    
	\end{subfigure}
	~
	\begin{subfigure}[b]{\myTextwidth\textwidth}                
		\centering
		\includegraphics[width=\myScale\linewidth]{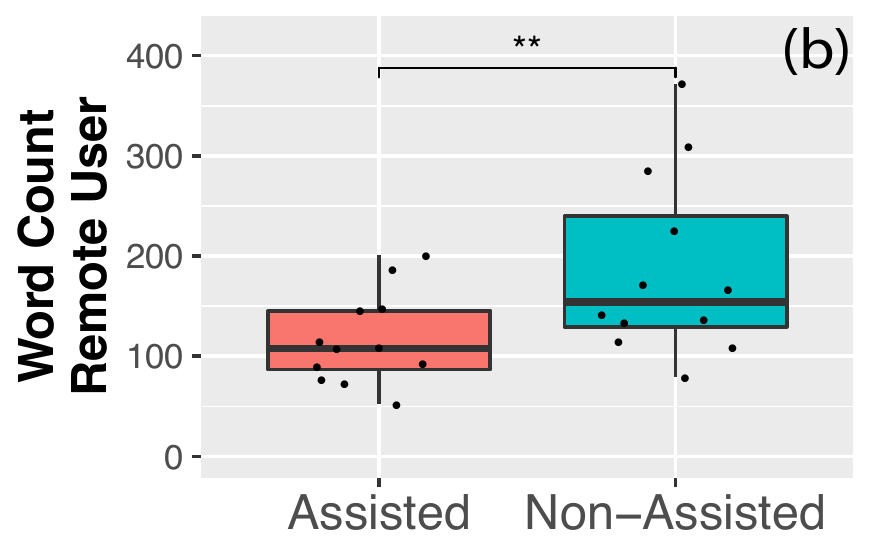}
		\label{fig:wordcountremoteuser}
	\end{subfigure}
	~
	\begin{subfigure}[b]{\myTextwidth\textwidth}
		\centering
		\includegraphics[width=\myScale\linewidth]{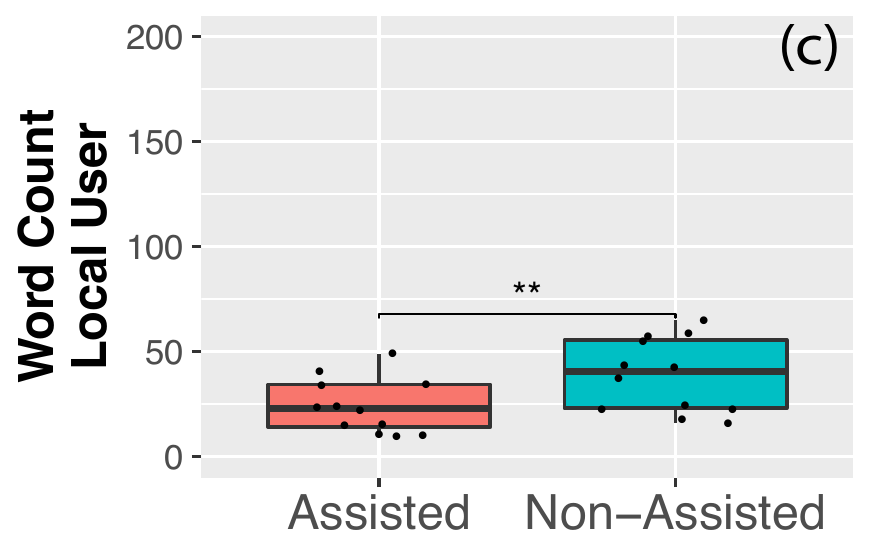}
		\label{fig:wordcountlocaluser}        
	\end{subfigure}
	~
	\begin{subfigure}[b]{\myTextwidth\textwidth}    
		\centering
		\includegraphics[width=\myScale\linewidth]{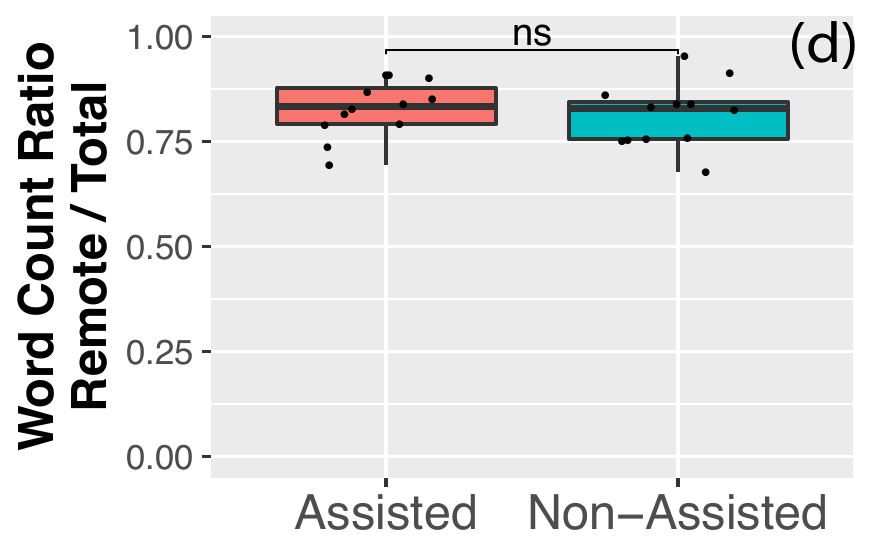}
		\label{fig:wordcountratio}
	\end{subfigure}
	~
	\begin{subfigure}[b]{\myTextwidth\textwidth}                
		\centering
		\includegraphics[width=\myScale\linewidth]{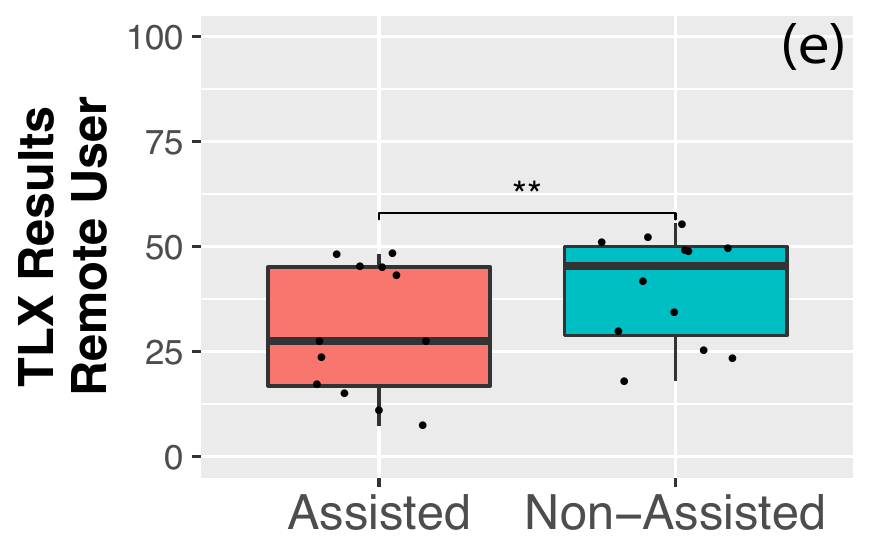}
		\label{fig:tlxresultsremoteuser}
	\end{subfigure}
	~
	\begin{subfigure}[b]{\myTextwidth\textwidth}
		\centering
		\includegraphics[width=\myScale\linewidth]{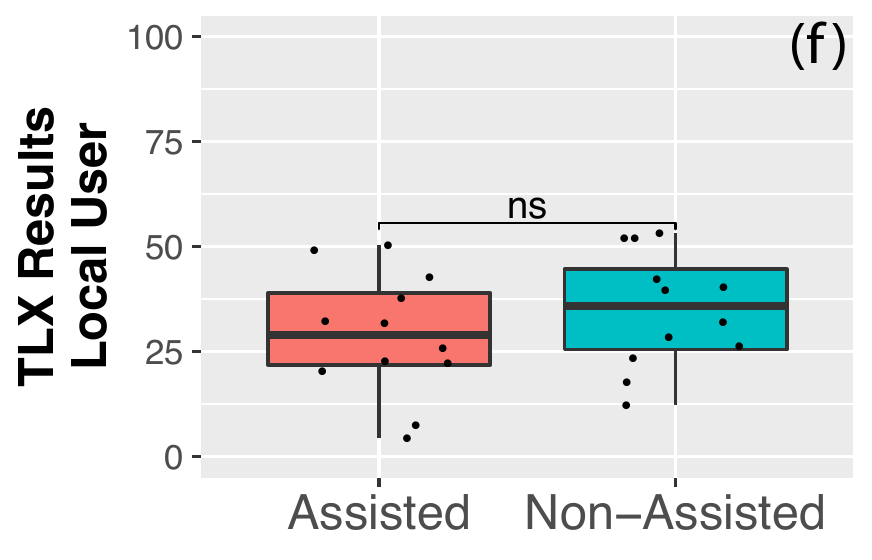}
		\label{fig:tlxresultslocaluser}
	\end{subfigure}
	~
	\begin{subfigure}[b]{\myTextwidth\textwidth}
		\centering
		\includegraphics[width=\myScale\linewidth]{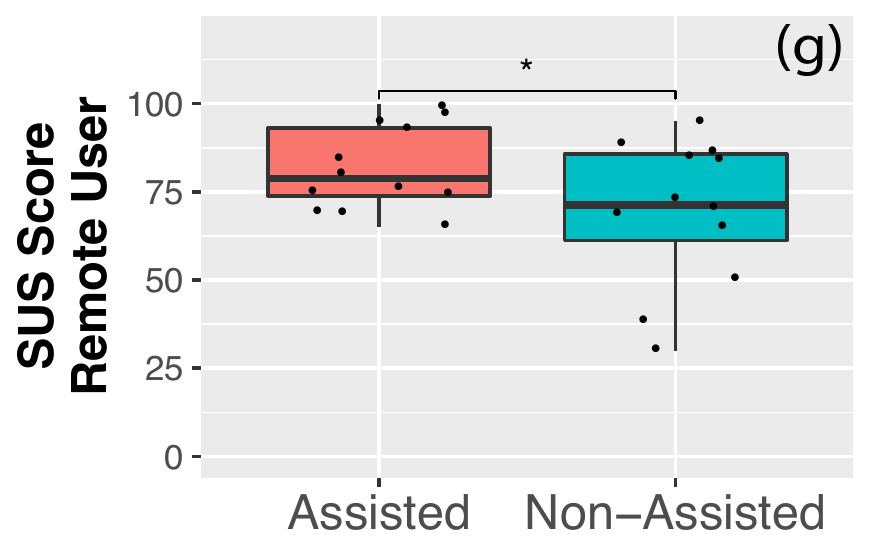}
		\label{fig:susscoreremoteuser}
	\end{subfigure}
	~
	\begin{subfigure}[b]{\myTextwidth\textwidth}    
		\centering
		\includegraphics[width=\myScale\linewidth]{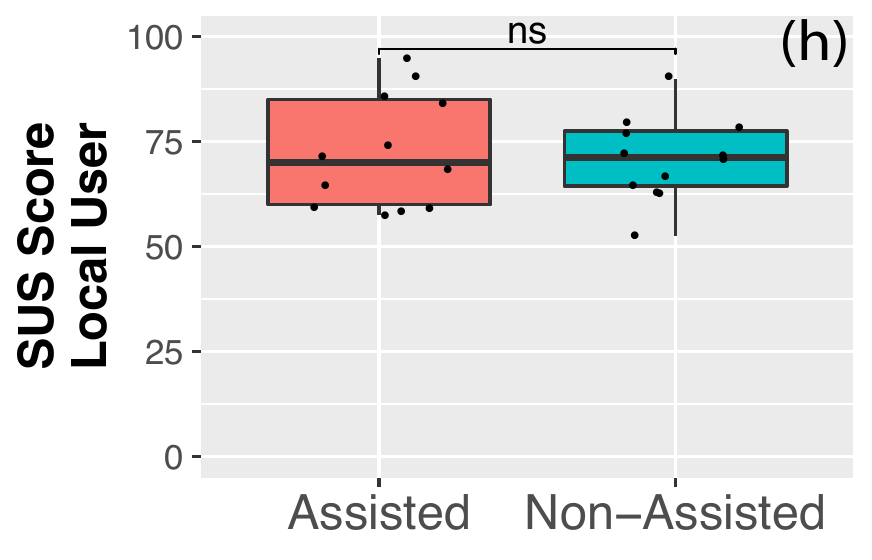}
		\label{fig:susscorelocaluser}
	\end{subfigure}
	\caption	
	{ \textbf{Comparative Analysis of Robot Conditions.}
		The Diagrams show the mean completion times, word counts, TLX and SUS scores for the assisted and non-assisted condition of the robot. Starred samples yield a significant difference due to the robot's condition without a significant interaction between the mode used and the remote user as a factor (cf. Table \ref{tab:remote-expert-ttest-results}). Levels of Significance: ns (not significant) $p \geq .050$; * $p < .050$; ** $p < .010$; *** $p < .001$; **** $p < .0001$}
	\label{fig:resultDiagram}
\end{figure*}

Concerning task performance, a significant ($p < .001$) decrease of mean completion time, from \SI{189.3}{s} to \SI{138.2}{s}, is observed when the robot is in assisted mode compared to the non-assisted mode (cf. Figure \ref{fig:resultDiagram}a). Hence, the teams performed the task 37\% faster in assisted mode. Furthermore, no significant correlation (as per Pearson \cite{benesty2009pearson}) between completion time and demographics was found for either of the participant groups, which is also true for the prior gaming experience.

Regarding the dialogue required for task coordination, from non-assisted to assisted mode there is a significant drop of word count for the remote users ($p = .005$) as well as for the local users ($p = .008$) leading to a total word count reduction of 38\% (cf. Figure \ref{fig:resultDiagram}b,c). The effect of the robot's condition on the word count ratio between remote and local user is non-significant.

In terms of participants' perceived workload, the robot's assistance significantly ($p = .001$) decreases from an average TLX score of 40 to 30 (lower is better), i.e. by 25\%. However, no significant effect ($p = .058$) is identified for the TLX score of the local user (cf. Figure \ref{fig:resultDiagram}e,f). In both conditions, the TLX ratings by the local users are close to the lower boundary of the scale.  

The overall system usability, as per the SUS questionnaire, is rated significantly higher ($p = .016$) by the remote users when the robot is in Assisted Mode. Per the TLX ratings, the robot's condition (i.e. Assisted/Non-Assisted Mode) has no significant ($p = .443$) effect on the SUS ratings for the local user group and they approach the upper boundary of the scale in both cases. The SUS results are summarised in Figure \ref{fig:resultDiagram}g,h.

We note that for the Assisted Mode, the perceived task load, as per TLX, was rated low, i.e. $\leq 30.0$ out of 100 (lower is better) for both user groups. Furthermore, the SUS ratings were high for both groups, i.e. with usability scores $\geq 72.5$ out of 100 (higher is better).



\subsection{Qualitative Analysis}

The major novelty of the proposed remote assistance scenario is the remote user's ability to physically interact with the workspace environment. This introduces new solution strategies and behaviours which are reflected in the collected video material. 

Across the different participant pairs, a general problem-solving strategy could be observed, which consists of four phases: scene exploration, spacial guidance, local task solving and retraction. During scene exploration, the remote user uses both camera views to diagnose a problem. Once the problem is identified, the local user is guided to a scene object through tooltip gestures and verbal instructions. The ratios between the usage of those two communication means vary strongly between remote users. While some participants gave many verbal instructions, others preferred the use of the tooltip for directions. Notably, in one instance, the remote user was able to guide the local user through the pipe checking task without using any verbal instructions. Instead, they used pointing gestures with the manipulator for navigation to a pipe that required checking and made the robot retreat to a crouch position to signal to the local user that the checking process was completed.

Another observation was that in this phase the remote user can in some instances control the motion speed of the local user by the amount of tip deflection. The bigger the angle with which the remote user steers the tip to one side, the higher the speed of the local user following it. This multimodal guiding strategy led to a more smooth interaction compared to ones where the remote user would solely use verbal instructions for navigation. 

After the object of interest is reached, there is a transition from shared control over the robot to remote user control as the local user holds the robot in place during manipulation. After a subtask is solved, the local user retreats away from the workspace to allow the remote user further exploration of the scene and the working cycle starts over again.


\section{Discussion}
\label{sec:remote_discussion}

In terms of task performance, we found that the handheld robot's autonomy and task knowledge contributes to a more efficient collaboration as it reduces the time to complete the task, which supports \ref{H1}. Regarding communication between the remote user and the local user, the qualitative results show that the remote control of the robot's tip extends the means of communication as it can be used for deictic gestures, such as pointing for navigation. 

\ref{H2} is partially supported as the robot's assistance features reduce the amount of verbal communication required to solve the task. We suggest that the reason for this observation is that the robot's aiming feature replaces the remote user's instructions for low scale motion. For example, in the non-assisted condition, remote users often relied on verbal instructions such as \textit{Go a bit more to the left, please} or  \textit{Up, up, up} during navigation to a task object. Whereas, when aiming was assisted, it was more clear to the local user where to go since they could just follow the tip's direction until reaching their aim. Furthermore, the robot's task knowledge accelerates processes which are otherwise interrupted through information requests. 

As the introduction of task knowledge in assisted mode changes the distribution of knowledge between the three interacting agents, we expected to observe a change in relative verbal communication amounts. However, this part of \ref{H2} is not supported by the results. Presumably, this is because in assisted mode the robot's task knowledge facilitates navigation, which cuts down required verbal commands by the remote user. At the same time, the robot executes local tasks autonomously, which reduces the number of instances where the local user has to report the system's current state. That way, the robot's assistance reduces the communication amount evenly so that the ratio won't change. 

In terms of the robot's effect on collaboration quality, we argue that a collaborative robot should be characterised by reducing workload and high usability for users. While this is supported by the overall TLX and SUS results for both users, improvement through the robot's assistance feature could only be found for remote users. This is less surprising, given that this work is focused on remote access and the assistive features were mainly designed to facilitate control of the robot. For example, the aiming feature takes away task load from the remote user while it does not make a notable difference for the local user whether the tip is controlled by the remote user or through the robot's assistance. Therefore, \ref{H3} and \ref{H4} are partially supported and we conclude that the assisted mode mainly benefits the work of the remote user.

The fact that the performance was independent of demographics and prior gaming experience, indicates that the system is suitable for people without technical expertise about the robot or the user interface. This makes the robot and the remote control system accessible for a broad range of applications, however, we suggest that different tasks require varying training approaches.

We note that setup described here is a first attempt of collaborative telepresence with two humans in the loop. As the robot is wired to the remote interface, the proposed testing setup does not take into account possible lags of the camera or time-delays in tip actuation, which might affect collaboration in a long-distance setup i.g. over the internet. This is because this study focused on the interaction experience of the two participants through the handheld robot. 

\section{Conclusion}
So far, remote collaboration and assistance between two humans has been limited to audio-visual instructions and feedback \cite{Gauglitz:2014hr,Gauglitz:2014gr,Bordegoni:iANORb1F}. At the same time, previous works on robot-based collaboration setups focused on specialised telemanipulation \cite{Bellamine:2002kt, Ballantyne:2002jc}
and mobile applications \cite{Haynes:2017jb}. The work presented in this paper explores a combination of both fields so that the resulting system benefits from the strengths of each. For remote assistance systems, that is the convenience of being able to rely on a human partner on the local work site while the telemanipulation aspect grants direct and immediate physical access to the helping expert.

We investigated the use of a handheld robot in our proposed collaborative remote assistance setup within a helper-worker scenario. 
We assessed the robot's capabilities through user studies, where a local user operates the handheld robot while being assisted by a remote user who gives verbal instructions and controls the robot's tip.

Regarding the research question \ref{Q} \textit{``How does the handheld robot's autonomy and task knowledge affect performance and communication in a remote assistance setup?''}, our studies show that a handheld robot can mediate task information and physical interaction in collaborative assistance. 
Importantly, the robot's partial autonomy improves task performance with respect to time efficiency, workload and required communication bandwidth.
Namely, the task was completed 37\% faster, remote user's work load decreased by 25\% and the required verbal communication by 38\%.
This means that when the robot was helping, the task was completed in a shorter time, while being easier to carry out for the remote user as less instructions were needed thanks to the delegation of subtasks to the robot. 


From our observations, enabling the operator to control the angle of the detail view camera, the direct physical interaction and the use of the spacial gestures proved effective to facilitate and in part replace wordy instructions and verbal information requests. This knowledge is useful to inform the design of future remote assistance systems.

The concept of object selection for task delegation to the robot is key to reduce the operator's cognitive task load, which we believe generalises to other telepresence setups. It might also help with problems that are introduced by time-delays. As the robot is autonomous for short durations, a fast feedback response might become less critical.


This work is a first attempt to evaluate handheld collaborative robots in a remote assistance scenario, a setup that can leverage current robot constraints such as incomplete task knowledge and motion competences with the already well-established communication technologies.

\subsection*{Acknowledgements}
Thanks to the German Academic Scholarship Foundation and UK's EPSRC for funding. Stated opinions are the authors' and not of the funders'.

\ifCLASSOPTIONcaptionsoff
  \newpage
\fi



%

\bibliographystyle{IEEEtran}
\bibliography{IEEEfull,references}

\begin{thebibliography}{10}
\providecommand{\url}[1]{#1}
\csname url@samestyle\endcsname
\providecommand{\newblock}{\relax}
\providecommand{\bibinfo}[2]{#2}
\providecommand{\BIBentrySTDinterwordspacing}{\spaceskip=0pt\relax}
\providecommand{\BIBentryALTinterwordstretchfactor}{4}
\providecommand{\BIBentryALTinterwordspacing}{\spaceskip=\fontdimen2\font plus
\BIBentryALTinterwordstretchfactor\fontdimen3\font minus
  \fontdimen4\font\relax}
\providecommand{\BIBforeignlanguage}[2]{{%
\expandafter\ifx\csname l@#1\endcsname\relax
\typeout{** WARNING: IEEEtran.bst: No hyphenation pattern has been}%
\typeout{** loaded for the language `#1'. Using the pattern for}%
\typeout{** the default language instead.}%
\else
\language=\csname l@#1\endcsname
\fi
#2}}
\providecommand{\BIBdecl}{\relax}
\BIBdecl

\bibitem{Gauglitz:2014hr}
S.~Gauglitz, B.~Nuernberger, M.~Turk, and T.~H{\"o}llerer, ``{In touch with the
  remote world: remote collaboration with augmented reality drawings and
  virtual navigation},'' in \emph{VRST '14 Proceedings of the 20th ACM
  Symposium on Virtual Reality Software and Technology}.\hskip 1em plus 0.5em
  minus 0.4em\relax New York, New York, USA: ACM Press, 2014, pp. 197--205.

\bibitem{Linn:ZL0DUdp_}
C.~Linn, S.~Bender, J.~Prosser, K.~Schmitt, and D.~Werth, ``{Virtual remote
  inspection{\textemdash}A new concept for virtual reality enhanced real-time
  maintenance},'' in \emph{2017 23rd International Conference on Virtual System
  {\&} Multimedia (VSMM)}, 2017.

\bibitem{Ballantyne:2002jc}
G.~H. Ballantyne, ``{Robotic surgery, telerobotic surgery, telepresence, and
  telementoring},'' \emph{Surgical Endoscopy}, vol.~16, no.~10, pp. 1389--1402,
  Oct. 2002.

\bibitem{Dell:2003tc}
J.~Ou, S.~R. Fussell, X.~Chen, L.~D. Setlock, and J.~Yang, ``{Gestural
  Communication over Video Stream: Supporting Multimodal Interaction for Remote
  Collaborative Physical Tasks},'' in \emph{ICMI '03: Proceedings of the 5th
  international conference on Multimodal interfaces}, 2003, pp. 242--249.

\bibitem{Gauglitz:2014gr}
S.~Gauglitz, B.~Nuernberger, M.~Turk, T.~H{\"o}llerer, and e.~design,
  ``{World-stabilized annotations and virtual scene navigation for remote
  collaboration},'' in \emph{the 27th annual ACM symposium}.\hskip 1em plus
  0.5em minus 0.4em\relax New York, New York, USA: ACM Press, 2014, pp.
  449--459.

\bibitem{Veronneau:2020dr}
C.~Veronneau, J.~Denis, L.-P. Lebel, M.~Denninger, V.~Blanchard, A.~Girard, and
  J.-S. Plante, ``{Multifunctional Remotely Actuated 3-DOF Supernumerary
  Robotic Arm Based on Magnetorheological Clutches and Hydrostatic Transmission
  Lines},'' \emph{IEEE Robotics and Automation Letters}, vol.~5, no.~2, pp.
  2546--2553, Feb. 2020.

\bibitem{Bordegoni:iANORb1F}
M.~Bordegoni, F.~Ferrise, E.~Carrabba, M.~Di~Donato, M.~Fiorentino, and A.~E.
  Uva, ``{An application based on Augmented Reality and mobile technology to
  support remote maintenance},'' in \emph{proceedings of 7th international
  Symposium on Artificial Intelligence, Robotics and Automation in Space,
  iSAIRAS}, 2014, pp. 131--135.

\bibitem{Wang:2014bt}
J.~Wang, Y.~Feng, C.~Zeng, and S.~Li, ``{An augmented reality based system for
  remote collaborative maintenance instruction of complex products},'' in
  \emph{2014 IEEE International Conference on Automation Science and
  Engineering (CASE)}.\hskip 1em plus 0.5em minus 0.4em\relax Taipei: IEEE,
  Jul. 2014, pp. 309--314.

\bibitem{Marturi:2017cm}
N.~Marturi, A.~Rastegarpanah, C.~Takahashi, M.~Adjigble, R.~Stolkin, S.~Zurek,
  M.~Kopicki, M.~Talha, J.~A. Kuo, and Y.~Bekiroglu, ``{Towards advanced
  robotic manipulation for nuclear decommissioning: A pilot study on
  tele-operation and autonomy},'' in \emph{2016 International Conference on
  Robotics and Automation for Humanitarian Applications (RAHA)}.\hskip 1em plus
  0.5em minus 0.4em\relax IEEE, 2016.

\bibitem{Hearst:1999vr}
M.~A. Hearst, ``{Mixed-initiative interaction},'' \emph{IEEE Intelligent
  Systems}, vol.~14, no.~5, pp. 14--23, Sep. 1999.

\bibitem{Kortenkamp:1997tc}
D.~Kortenkamp, R.~P. Bonasso, D.~Ryan, and D.~Schreckenghost, ``{Traded Control
  with Autonomous Robots as Mixed Initiative Interaction},'' in \emph{AAAI
  Symposium on Mixed Initiative Interaction}, 1997, pp. 89--94.

\bibitem{GreggSmith:2015bh}
A.~Gregg-Smith and W.~W. Mayol-Cuevas, ``{The design and evaluation of a
  cooperative handheld robot},'' in \emph{2015 IEEE International Conference on
  Robotics and Automation (ICRA)}.\hskip 1em plus 0.5em minus 0.4em\relax IEEE,
  2015, pp. 1968--1975.

\bibitem{GreggSmith:2016cz}
------, ``{Inverse Kinematics and Design of a Novel 6-DoF Handheld Robot
  Arm},'' in \emph{2016 IEEE International Conference on Robotics and
  Automation (ICRA)}.\hskip 1em plus 0.5em minus 0.4em\relax IEEE, 2016, pp.
  2102--2109.

\bibitem{GreggSmith:2016hn}
------, ``{Investigating spatial guidance for a cooperative handheld robot},''
  in \emph{2016 IEEE International Conference on Robotics and Automation
  (ICRA)}.\hskip 1em plus 0.5em minus 0.4em\relax IEEE, 2016, pp. 3367--3374.

\bibitem{Stolzenwald:2018un}
J.~Stolzenwald and W.~Mayol-Cuevas, ``{I Can See Your Aim: Estimating User
  Attention From Gaze For Handheld Robot Collaboration},'' in \emph{2018
  IEEE/RSJ International Conference on Intelligent Robots and Systems (IROS)},
  Oct. 2018, pp. 3897--3904.

\bibitem{Stolzenwald:2019wi}
J.~Stolzenwald and W.~W. Mayol-Cuevas, ``{Rebellion and Obedience: The Effects
  of Intention Prediction in Cooperative Handheld Robots},'' in \emph{2019
  IEEE/RSJ International Conference on Intelligent Robots and Systems (IROS)},
  Macau, China, Nov. 2019, pp. 3012--3019.

\bibitem{Elsdon:2017is}
J.~Elsdon and Y.~Demiris, ``{Assisted painting of 3D structures using shared
  control with a hand-held robot},'' in \emph{2017 IEEE International
  Conference on Robotics and Automation (ICRA)}, 2017, pp. 4891--4897.

\bibitem{Elsdon:2018kl}
------, ``{Augmented Reality for Feedback in a Shared Control Spraying Task},''
  in \emph{2018 IEEE International Conference on Robotics and Automation
  (ICRA)}.\hskip 1em plus 0.5em minus 0.4em\relax IEEE, 2018, pp. 1939--1946.

\bibitem{Echtler:2003uo}
F.~Echtler, F.~Sturm, K.~Kindermann, G.~Klinker, J.~Stilla, J.~Trilk, and
  H.~Najafi, ``{The Intelligent Welding Gun: Augmented Reality for Experimental
  Vehicle Construction},'' in \emph{Virtual and Augmented Reality Applications
  in Manufacturing}.\hskip 1em plus 0.5em minus 0.4em\relax Springer London,
  2003, pp. 333--360.

\bibitem{Moravec:1988un}
H.~Moravec, ``{Mind children},'' Harvard University Press, Cambridge, MA, 1988.

\bibitem{AbiFarraj:2017ex}
F.~Abi-Farraj, T.~Osa, N.~P.~J. Peters, G.~Neumann, and P.~R. Giordano, ``{A
  learning-based shared control architecture for interactive task execution},''
  \emph{2017 IEEE International Conference on Robotics and Automation (ICRA)},
  pp. 329--335, May 2017.

\bibitem{Kritzler:2016ie}
M.~Kritzler, M.~Murr, and F.~Michahelles, ``{RemoteBob - Support of On-site
  Workers via a Telepresence Remote Expert System},'' in \emph{the 6th
  international conference}.\hskip 1em plus 0.5em minus 0.4em\relax New York,
  New York, USA: ACM Press, 2016, pp. 7--14.

\bibitem{Sodhi:JJe-ACW9}
R.~S. Sodhi, B.~R. Jones, D.~Forsyth, B.~P. Bailey, and G.~Maciocci,
  ``{BeThere: 3D mobile collaboration with spatial input},'' in \emph{CHI
  Changing Perspectives}, Paris, France, 2013, pp. 179--188.

\bibitem{KostiaRobert:2013un}
K.~Robert, D.~Zhu, W.~Huang, and L.~Alem, ``{MobileHelper: Remote Guiding Using
  Smart Mobile Devices, Hand Gestures and Augmented Reality},'' in
  \emph{SIGGRAPH Asia Symposium on Mobile Graphics and Interactive
  Applications. ACM}, Aug. 2013.

\bibitem{Bottecchia:2010co}
S.~Bottecchia, J.-M. Cieutat, and J.-P. Jessel, ``{T.A.C: Augmented Reality
  System for Collaborative Tele-Assistance in the Field of Maintenance through
  Internet},'' in \emph{Proceedings of the 1st Augmented Human International
  Conference}.\hskip 1em plus 0.5em minus 0.4em\relax New York, New York, USA:
  ACM Press, 2010.

\bibitem{Kuzuoka:DjdzJkGx}
H.~Kuzuoka and J.~Kosaka, ``{GestureMan PS: Effect of a head and a pointing
  stick on robot mediated communication},'' in \emph{Proceedings of HCII},
  2003, pp. 1416--1420.

\bibitem{Anonymous:pTXsTX3C}
H.~Kuzuoka, S.~Oyama, K.~Yamazaki, and M.~Mitsuishi, ``{GestureMan: a mobile
  robot that embodies a remote instructor's actions},'' in \emph{Proceedings of
  the ACM conference on Computer supported cooperative work}, 2000, pp.
  155--162.

\bibitem{Kurata:qhDyKTAw}
T.~Kurata, N.~Sakata, M.~Kourogi, H.~Kuzuoka, and M.~Billinghurst, ``{Remote
  collaboration using a shoulder-worn active camera/laser},'' in \emph{Eighth
  international symposium on wearable computers}, 2004, pp. 62--69.

\bibitem{Kangas:2018ky}
J.~Kangas, A.~Sand, T.~Jokela, P.~Piippo, P.~Eskolin, M.~Salmimaa, and
  R.~Raisamo, ``{Remote Expert for Assistance in a Physical Operational
  Task},'' in \emph{Extended Abstracts of the 2018 CHI Conference}.\hskip 1em
  plus 0.5em minus 0.4em\relax New York, New York, USA: ACM Press, 2018.

\bibitem{Yamamoto:2018gt}
T.~Yamamoto, M.~Otsuki, H.~Kuzuoka, and Y.~Suzuki, ``{Tele-Guidance System to
  Support Anticipation during Communication},'' \emph{Multimodal Technologies
  and Interaction}, vol.~2, no.~3, p.~55, Sep. 2018.

\bibitem{Bellamine:2004gd}
M.~Bellamine, N.~Abe, K.~Tanaka, P.~Chen, and H.~Taki, ``{A Virtual Reality
  Based System for Remote Maintenance of Rotating Machinery},'' in
  \emph{Embedded and Ubiquitous Computing}.\hskip 1em plus 0.5em minus
  0.4em\relax Springer, Berlin, Heidelberg, Aug. 2004, pp. 164--173.

\bibitem{Bellamine:2002kt}
M.~Bellamine, N.~Abe, K.~Tanaka, and H.~Taki, ``{Remote machinery maintenance
  system with the use of virtual reality},'' in \emph{Proceedings of the 2003
  IEEE Computer Society Conference on Computer Vision and Pattern Recognition
  (CVPR{\textquoteright}03)}, 2002, pp. 38--43.

\bibitem{Schiele:2003tk}
A.~Schiele and G.~Visentin, ``{The ESA human arm exoskeleton for space robotics
  telepresence},'' in \emph{proceedings of 7th international Symposium on
  Artificial Intelligence, Robotics and Automation in Space, iSAIRAS}, Nara,
  Japan, 2003, pp. 19--23.

\bibitem{AbiFarraj:2016ec}
F.~Abi-Farraj, N.~Pedemonte, and P.~R. Giordano, ``{A visual-based shared
  control architecture for remote telemanipulation},'' in \emph{2016 IEEE/RSJ
  International Conference on Intelligent Robots and Systems (IROS)}.\hskip 1em
  plus 0.5em minus 0.4em\relax IEEE, Nov. 2016, pp. 4266--4273.

\bibitem{Haynes:2017jb}
G.~C. Haynes, D.~Stager, A.~Stentz, J.~M. Vande~Weghe, B.~Zajac, H.~Herman,
  A.~Kelly, E.~Meyhofer, D.~Anderson, D.~Bennington, J.~Brindza,
  D.~Butterworth, C.~Dellin, M.~George, J.~Gonzalez-Mora, M.~Jones, P.~Kini,
  M.~Laverne, N.~Letwin, E.~Perko, C.~Pinkston, D.~Rice, J.~Scheifflee,
  K.~Strabala, M.~Waldbaum, and R.~Warner, ``{Developing a Robust Disaster
  Response Robot: CHIMP~and the Robotics Challenge},'' \emph{Journal of Field
  Robotics}, vol.~34, no.~2, pp. 281--304, Feb. 2017.

\bibitem{Imaida:2004bm}
T.~Imaida, Y.~Yokokohji, T.~Doi, M.~Oda, and T.~Yoshikawa,
  ``{Ground{\textendash}Space Bilateral Teleoperation of ETS-VII Robot Arm by
  Direct Bilateral Coupling Under 7-s Time Delay Condition},'' \emph{IEEE
  Transactions on Robotics and Automation}, vol.~20, no.~3, pp. 499--511, Jun.
  2004.

\bibitem{Wedler:2018vn}
A.~Wedler, M.~Wildee, J.~Reilla, M.~J. Schustera, M.~Vayugundlaa, S.~G.
  Brunnera, K.~Bussmanna, A.~D{\"o}mela, M.~Drauschkea, H.~Gmeinera,
  H.~Lehnera, P.~Lehnera, M.~G. M{\"u}llera, W.~St{\"u}rzla, R.~Triebela,
  B.~Vodermayera, A.~B{\"o}rnerb, R.~Krennc, A.~Dammannd, U.-C. Fiebigd,
  E.~Staudingerd, F.~Wenzh{\"o}fere, S.~Fl{\"o}gelf, S.~Sommerf, T.~Asfourg,
  M.~Fladh, S.~Hohmannh, M.~Brandaueri, and A.~O. Albu-Sch{\"a}ffera, ``{From
  single autonomous robots toward cooperative robotic interactions for future
  planetary exploration missions },'' in \emph{Preceedings of the 69th
  International Astronautical Congress IAC}, Bremen, Sep. 2018.

\bibitem{Anonymous:XQistSCe}
{Handheld Robotics: handheldrobotics.org}.

\bibitem{Wilson:2014el}
T.~G. Wilson, ``{Advancement of Technology and Its Impact on Urologists:
  Release of the da Vinci Xi, A New Surgical Robot},'' \emph{European Urology},
  vol.~66, no.~5, pp. 793--794, Nov. 2014.

\bibitem{Skilton:2018gw}
R.~Skilton, N.~Hamilton, R.~Howell, C.~Lamb, and J.~Rodriguez, ``{MASCOT 6:
  Achieving high dexterity tele-manipulation with a modern architectural design
  for fusion remote maintenance},'' \emph{Fusion Engineering and Design}, vol.
  136, pp. 575--578, Nov. 2018.

\bibitem{Troccaz:1996ck}
J.~Troccaz and Y.~Delnondedieu, ``{Semi-active guiding systems in surgery. A
  two-dof prototype of the passive arm with dynamic constraints (PADyC)},''
  \emph{Mechatronics}, vol.~6, no.~4, pp. 399--421, Jun. 1996.

\bibitem{Nielsen:2008vu}
C.~W. Nielsen, D.~J. Bruemmer, and D.~A. Few, ``{Framing and evaluating
  human-robot interactions},'' in \emph{Proceedings of the Workshop on Metrics
  for Human-Robot Interaction}, 2008, pp. 29--36.

\bibitem{Hart:1988ho}
S.~G. Hart and L.~E. Staveland, ``{Development of NASA-TLX (Task Load Index):
  Results of Empirical and Theoretical Research},'' in \emph{Human Mental
  Workload}.\hskip 1em plus 0.5em minus 0.4em\relax Elsevier, 1988, pp.
  139--183.

\bibitem{brooke1996quick}
J.~Brooke, \emph{{"SUS - A quick and dirty usability scale." Usability
  Evaluation In Industry}}.\hskip 1em plus 0.5em minus 0.4em\relax CRC Press,
  Jun. 1996.

\bibitem{Hart:1988hoa}
S.~G. Hart and L.~E. Staveland, ``{Development of NASA-TLX (Task Load Index):
  Results of Empirical and Theoretical Research},'' \emph{Advances in
  psychology}, vol.~52, pp. 139--183, Jan. 1988.

\bibitem{Adamides:2017jf}
G.~Adamides, C.~Katsanos, Y.~Parmet, G.~Christou, M.~Xenos, T.~Hadzilacos, and
  Y.~Edan, ``{HRI usability evaluation of interaction modes for a teleoperated
  agricultural robotic sprayer},'' \emph{Applied Ergonomics}, vol.~62, pp.
  237--246, Jul. 2017.

\bibitem{benesty2009pearson}
J.~Benesty, J.~Chen, Y.~Huang, and I.~Cohen, ``{Pearson correlation
  coefficient},'' in \emph{Noise reduction in speech processing}.\hskip 1em
  plus 0.5em minus 0.4em\relax Springer, 2009.

\end{thebibliography}

%
%
%
%
%


\vfill


\end{document}